%% file: egpaper_final.tex
\documentclass[runningheads]{llncs}
\input{macros}
\usepackage{graphicx}
\usepackage{amsmath,amssymb} % define this before the line numbering.
\usepackage{color}
\usepackage{times}
\usepackage{color}
\usepackage{balance}
\usepackage{subcaption}
\usepackage{xcolor}

\DeclareMathOperator*{\argmax}{argmax}
\DeclareMathOperator*{\argmin}{argmin}
\usepackage[width=122mm,left=12mm,paperwidth=146mm,height=193mm,top=12mm,paperheight=217mm]{geometry}
\begin{document}
% \renewcommand\thelinenumber{\color[rgb]{0.2,0.5,0.8}\normalfont\sffa~ mily\scriptsize\arabic{linenumber}\color[rgb]{0,0,0}}
% \renewcommand\makeLineNumber {\hss\thelinenumber\ \hspace{6mm} \rlap{\hskip\textwidth\ \hspace{6.5mm}\thelinenumber}}
% \linenumbers
\pagestyle{headings}
\mainmatter
\def\ECCV18SubNumber{2849}  % Insert your submission number here

\title{ConvNets and ImageNet Beyond Accuracy:
Understanding Mistakes and Uncovering Biases}

\titlerunning{ImageNet Beyond Accuracy}

\authorrunning{Stock and Cisse}

\author{Pierre Stock\thanks{Mail: pstock@fb.com} and Moustapha Cisse\thanks{Now Google AI Ghana Lead. Mail: moustaphacisse@google.com}}
\institute{Facebook AI Research}

%%%%%%%%% TITLE
%\title{ConvNets and ImageNet Beyond Accuracy: \\ Explanations, Bias Detection, Adversarial Examples and  Model Criticism}

%\author{Pierre Stock\\
%Facebook AI Research\\
%Institution1 address\\
%{\tt\small pstock@fb.com}
% For a paper whose authors are all at the same institution,
% omit the following lines up until the closing ``}''.
% Additional authors and addresses can be added with ``\and'',
% just like the second author.
% To save space, use either the email address or home page, not both
%\and
%Moustapha Cisse\\
%Facebook AI Research\\
%First line of institution2 address\\
%{\tt\small moustaphacisse@fb.com}
%}

\maketitle
%\thispagestyle{empty}

%%%%%%%%% ABSTRACT
\begin{abstract}
ConvNets and ImageNet have driven the recent success of deep learning for image classification. However, the marked slowdown in performance improvement combined with the lack of robustness of neural networks to adversarial examples and their tendency to exhibit undesirable biases question the reliability of these methods. This work investigates these questions from the perspective of the end-user by using human subject studies and explanations. The contribution of this study is threefold. We first experimentally demonstrate that the accuracy and robustness of ConvNets measured on ImageNet are vastly underestimated. Next, we show that explanations can mitigate the impact of misclassified adversarial examples from the perspective of the end-user. We finally introduce a novel tool for uncovering the undesirable biases learned by a model. These contributions also show that explanations are a valuable tool both for improving our understanding of ConvNets' predictions and for designing more reliable models.%\footnote{corresponding author: moustaphacisse@fb.com}.
\end{abstract}

\section{Introduction}

    Convolutional neural networks~\cite{lecun1998gradient,krizhevsky2012imageNet} and ImageNet~\cite{deng2009imagenet} (the dataset and the challenge) have been instrumental to the recent breakthroughs in computer vision. ImageNet has provided ConvNets with the data they needed to demonstrate their superiority compared to the previously used handcrafted features such as Fisher Vectors~\cite{perronnin2010improving}. In turn, this success has triggered a renewed interest in convolutional approaches. Consequently, novel architectures such as ResNets~\cite{he2016deep} and DenseNets~\cite{huang2016densely} have been introduced to improve the state of the art performance on ImageNet. The impact of this virtuous circle has permeated all aspects of computer vision and deep learning at large. Indeed, the use of feature extractors pre-trained on ImageNet is now ubiquitous. For example, the state of the art image segmentation~\cite{girshick2015fast,he2017mask} or pose estimation models~\cite{insafutdinov2016deepercut,bulat2016human} heavily rely on pre-trained ImageNet features. Besides, convolutional architectures initially developed for image classification such as Residual Networks are now routinely used for machine translation~\cite{gehring2017convolutional} and speech recognition~\cite{wang2017residual}.

\begin{figure}
\centering
\includegraphics[width=0.5\linewidth]{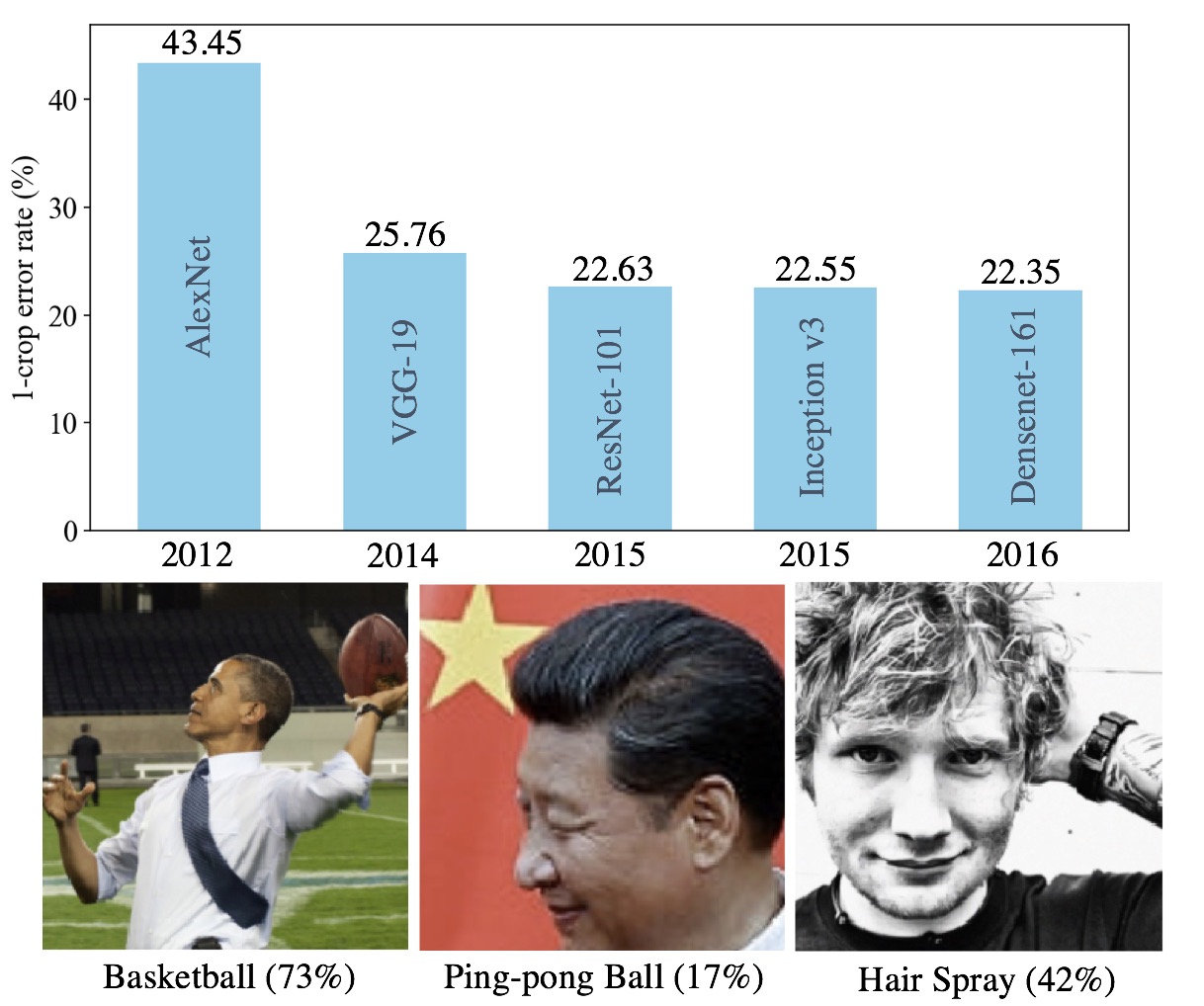}
   \caption{\textbf{Top:} Performance evolution of various CNN architectures on ImageNet. \textbf{Bottom:} Some images sampled from the Internet and misclassified by a ResNet-101.}
\label{fig:long}
\label{imagenet_cherry}
\end{figure}

  Since 2012, the top-1 error of state of the art (SOTA) models on ImageNet has been reduced from $43.45\%$ to $22.35\%$. Recently, the evolution of the best performance seems to plateau (see Figure \ref{imagenet_cherry}) despite the efforts in designing novel architectures~\cite{he2016deep,huang2016densely}, introducing new data augmentation schemes~\cite{zhang2017mixup} and optimization algorithms~\cite{kingma2014adam}. Concomitantly, several studies have demonstrated the lack of robustness of deep neural networks to adversarial examples~\cite{szegedy2013intriguing,goodfellow2014explaining,cisse2017houdini} and raised questions about their tendency to exhibit (undesirable) biases~\cite{ritter2017cognitive}. Adversarial examples~\cite{szegedy2013intriguing} are synthetic images designed to be indistinguishable from natural ones by a human, yet they are capable of fooling the best image classification systems. Undesirable biases are patterns or behaviors learned from the data, they are often highly influential in the decision of the model but are not aligned with the values of the society in which the model operates. Examples of such biases include \emph{racial} and \emph{gender biases}~\cite{bolukbasi2016man}.
While accuracy has been the leading factor in the broad adoption of deep learning across the industries, its \emph{sustained improvement} together with other desirable properties such as \emph{robustness to adversarial examples} and \emph{immunity to biases} will be critical in maintaining the trust in the technology. It is therefore essential to improve our understanding of these questions from the perspective of the end-user in the context of ImageNet classification. %(especially ImageNet, the primary image classification benchmark).
%The lack of \emph{robustness}, the \emph{biases} and the marked \emph{slowdown in performance improvement} can be due to several constituents such as the training dataset, the inductive bias (\textit{e.g.} empirical risk minimization), the optimization algorithm or the architecture of the models.
%While it can be difficult to identify the contribution of each of these factors in isolation (through ablation studies for example), the predictions of a state of the art convolutional network trained on ImageNet contains valuable information which can lead to a better understand these phenomena, provided one has the right tools to analyze them.

In this work, we take a step in this direction by assessing the predictions of SOTA models on the validation set of ImageNet.
%we conduct several human studies exploiting the predictions of SOTA models and explore various techniques to analyze the predictions of the model.
We show that \emph{human studies} and \emph{explanations} can be valuable tools to perform this task. Human studies yield a new judgment of the quality of the model's predictions from the perspective of the end-user, in addition to the traditional ground truth used to evaluate the model. We also use both \emph{feature-based} and \emph{example-based} explanations. On the one hand, explaining the prediction of a black box classifier by a subset of the features of its input can yield valuable insights into the workings of the models and underline the essential features in its decision. On the other hand, example-based explanations provide an increased \emph{interpretability} of the model by highlighting instances representative of the distribution of a given category as captured by the model. The particular form of example-based explanation we use is called \emph{model criticism}~\cite{kim2016examples}. It combines both \emph{prototypes} and \emph{criticisms} and is proven to better capture the complex distributions of natural images. Therefore, it facilitates human understanding.
Our main findings are summarized below:
\begin{itemize}
\item The accuracy of convolutional networks evaluated on ImageNet is \emph{vastly underestimated}. We find that  when the mistakes of the model are assessed by human subjects and considered correct when at least four out of five humans agree with the model's prediction, the top-1 error of a ResNet-101 trained on Imagenet and evaluated on the standard validation set decreases from $22.69\%$ to \textbf{$9.47\%$}. Similarly, the top-5 error decreases from $6.44\%$ to \textbf{$1.94\%$}. This observation holds across models. It explains the marked slowndown in accuracy improvement and suggests that ImageNet is almost solved.
\item The robustness of ConvNets to adversarial examples is also underestimated. In addition, we show that providing explanations helps to mitigate the misclassification of adversarial examples from the perspective of the end-user.
\item \emph{Model Criticism} is a valuable tool for detecting the biases in the predictions of the models. Further, adversarial examples can be effectively used for model criticism.
\end{itemize}

  Similar observations to our first point existed in prior work~\cite{blog}. However, the scale, the conclusions and the implications of our study are different. Indeed we consider that if top-5 error is the measure of interest, ImageNet is (almost) solved. In the next section, we summarize the related work before presenting our experiments and results in details.
%\newpage
\section{Related Work}

\paragraph{Adversarial examples.}
\label{sec:advex}

Deep neural networks can achieve high accuracy on previously unseen examples while being vulnerable to small adversarial perturbations of their inputs~\cite{szegedy2013intriguing}. Such perturbed inputs, called adversarial examples, have recently aroused keen interest in the community~\cite{goodfellow2015explaining,szegedy2013intriguing,tabacof2015exploring,cisse2017houdini}. Several studies have subsequently analyzed the phenomenon~\cite{fawzi2015analysis,shaham2015understanding,fawzi2016robustness} and various approaches have been proposed to improve the robustness of neural networks~\cite{papernot2016distillation,cisse2017parseval,tramer2017ensemble,zhang2017mixup,guo2017countering}.
More closely related to our work are the different proposals aiming at generating better adversarial examples~\cite{goodfellow2015explaining,moosavi2015deepfool}.
Given an input (train or test) example $(\bx, \by)$, an adversarial
example is a perturbed version of the original pattern $\adv{\bx} = \bx +
\perturb{\bx}$ where $\perturb{\bx}$ is small enough for
$\adv{\bx}$ to be undistinguishable from $\bx$ by a human, but causes the network to predict an
incorrect target. Given the network $g_\theta$ (where $\theta$ is the set of parameters) and a $p$-norm, the adversarial example is formally defined as:
\begin{equation}
\label{eq:adv}
\bxprime = \argmax_{\bxprime:\|{\bxprime-\bx}\|_p\leq \adveps}
\loss\big(\network_\theta(\bxprime), \by\big)
\end{equation}
where $\adveps$ represents the strength of the adversary. Assuming the loss function $\loss(\cdot)$ is differentiable,
\cite{shaham2015understanding} propose to take the first
order taylor expansion of $\bx\mapsto\loss(\network_\theta(\bx),
\by)$ to compute $\perturb{\bx}$ by solving the following simpler problem:
\begin{equation}
\label{eq:fgsm}
\bxprime = \argmax_{\bxprime:\|{\bxprime-\bx}\|_p\leq \adveps}
\big(\nabla_{\bx} \loss(\network_\theta(\bx),
\by)\big)^T (\bxprime-\bx).
\end{equation}
When $p=\infty$, then $\bxprime = \bx + \adveps\cdot\text{sign}(\nabla_{\bx}
\loss(\network_\theta(\bx), \by))$ which corresponds to the \emph{fast gradient
  sign method}~\cite{goodfellow2015explaining}. If instead $p=2$, we obtain $\bxprime = \bx + \adveps\cdot
\nabla_{\bx} \loss(\network_\theta(\bx), \by)$ where $\nabla_{\bx} \loss(\network_\theta(\bx), \by)$ is often normalized. Optionally, one can perform more iterations of these steps using a smaller step-size. This strategy has several variants~\cite{moosavi2015deepfool,kurakin2016adversarial}. In the rest of the paper, we refer to this method by \emph{iterative fast gradient method} (IFGM) and will use it both to measure the robustness of a given model and to perform model criticism.

\paragraph{Model Criticism.}
Example-based explanations are a well-known tool in the realm of cognitive science for facilitating human understanding~\cite{Simon1972HumanPS}.  They have extensively been used in case-based reasoning (CBR) to improve the interpretability of the models~\cite{Aamodt1994CaseBasedRF,Bichindaritz2006CasebasedRI,Kim2014TheBC}. In most cases, it consists in helping a human to understand a complex distribution (or the statistics captured by a model) by presenting her with a set of prototypical examples. However, when the distribution or the model for which one is seeking explanation is complex (as is often the case with real-world data), prototypes may not be enough. Recently, Kim et al.~\cite{kim2016examples} have proposed to use, in addition to the prototypes, data points sampled from regions of the input space not well captured by the model or the prototypes. Such examples, called \emph{criticism}, are known to improve the human's mental model of the distribution.

Kim et al~\cite{kim2016examples} introduced MMD-critic, an approach inspired by bayesian model criticism~\cite{Gelman2006BayesianDA} to select the \emph{prototypes} and the \emph{critics} among a given set of examples. MMD-critic uses the maximum mean discrepancy~\cite{Gretton2006AKM} and large-scale submodular optimization~\cite{Badanidiyuru2014StreamingSM}. Given a set of examples $\mathcal{D} = \{(\bx, \by)\}_{i=1}^n$, let $S\subset \{1, \ldots,n\}$ such that $\mathcal{D}_S = \{\bx_i, i \in S\}$. Given a RKHS with kernel function $k(\cdot, \cdot)$, the prototypes of MMD-critic are selected by minimizing the maximum mean discrepancy between $\mathcal{D}$ and $\mathcal{D}_S$. This is formally written as:
\begin{equation}
\max_{S \in 2^{[n]}, |S|\leq m} J_b(S) =  \frac{2}{n|S|}\sum_{i \in [n], j\in S} k(x_i, x_j) - \frac{1}{|S|^2}\sum_{i,j\in S} k(x_i, x_j).
\end{equation}
%where $J_b(S)$ can be expressed as follows:
%\begin{equation*}
%J_b(S) = \frac{2}{n|S|}\sum_{i \in [n], j\in S} k(x_i, x_j) - \frac{1}{|S|^2}\sum_{i,j\in S} k(x_i, x_j)
%\end{equation*}
Given a set of prototypes, the criticisms are similarly selected using MMD to maximize the deviation from the prototypes. The objective function in this case is regularized to promote diversity among the criticisms. A greedy algorithm can be used to select both prototypes and criticisms since the corresponding optimization problems are provably submodular and monotone under certain conditions~\cite{Badanidiyuru2014StreamingSM}. In our experimental study, we will use MMD-critic as a baseline for example-based explanations.

\paragraph{Feature-based Explanation.}
\label{sec:lime}
  Machine learning and especially Deep Neural Networks (DNNs) lie at the core of more and more technological advances across various fields. However, those models are still widely considered as \textit{black boxes}, leading end users to mistrust the predictions or even the underlying models. In order to promote the adoption of such algorithms and to foster their positive technological impact, recent studies have been focusing on understanding a model from the human perspective \cite{lipton2016mythos,samek2017explainable,montavon2017methods,dong2017towards}.

In particular, \cite{lime} propose to explain the predictions of any classifier $g_{\theta}$ (denoted as $g$) by approximating it locally with an interpretable model $h$. The role of $h$ is to provide qualitative understanding between the input $x$ and the classifier's output $g(x)$ for a given class. In the case where the input $x$ is an image, $h$ will act on vector $x' \in \{0,1\}^d$ denoting the presence or absence of the $d$ super-pixels that partition the image $x$ to \textit{explain} the classifier's decision.

Finding the best explanation $\xi(x)$ among the candidates $h$ can be formulated as:
\begin{equation}
\xi(x) = \argmin_{h \in H} \mathcal L(g, h,\pi_x) + \Omega(h)
\end{equation}
where the best explanation minimizes a local weighted loss $\mathcal L$ between $g$ and $h$ in the vicinity $\pi_x$ of $x$, regularized by the complexity $\Omega(h)$ of such an explanation. The authors restrict $h \in H$ to be a linear model such that $h(z') = w_hz'$. They further define the vicinity of two samples using the exponential kernel
\begin{equation}
\pi_x(z) = e^{\frac{\|x-z\|^2}{\sigma ^2}}
\end{equation}
and define the local weighted loss $\mathcal L$ as:
\begin{equation}
\mathcal L(g, h, \pi_x) = \sum_{z, z'\in \mathcal Z}\pi_x(z)\left(g(z) - h(z')\right)^2
\end{equation}
where $\mathcal Z$ if the dataset of $n$ perturbed samples obtained from $x$ by randomly activating or deactivating some super-pixels in $x$. Note that $z'$ denotes the one-hot encoding of the super-pixels whereas $z$ is the actual image formed by those super-pixels. Finally, the interpretability of the representation is controlled by
\begin{equation}
\Omega(h) = \begin{cases}
      0       & \textrm{if}~~~\|w_h\|_0 < K \\
      +\infty & \textrm{otherwise} \\
   \end{cases}
\end{equation}
The authors solve this optimization problem by first selecting $K$ features with Lasso and then learning the weights $w_g$ \textit{via} least squares. In the case of DNNs, an explanation generated by this algorithm called LIME allows the user to just highlight the super-pixels with positive weights towards a specific class (see Fig~\ref{jeep}). In what follows we set $\sigma = 0.25$, $n=1000$ and and keep $K$ constant. In the following we refer to an image prompted solely with its top super-pixels as an image \textit{with attention}.

%%%%%%%%% BODY TEXT
\section{Experiments}

%-------------------------------------------------------------------------
\subsection{Human Subject Study of Classification Errors}

We conduct a study of the misclassifications of various pre-trained architectures (\textit{e.g.} ResNet-18, ResNet-101, DenseNet-121, DenseNet-161) with human subjects on Amazon Mechanical Turk (AMT).  To this end, we use the center-cropped images of size $224\times 224$ from the standard validation set of ImageNet.
The same setting holds in all our experiments. For every architecture, we consider all the examples misclassified by the model. Each of these examples is presented to five (5) different turkers together with the class (C) predicted by the model. Each turker is then asked the following question "\emph{Is class C relevant for this image}." The possible answers are \emph{yes} and \emph{no}. The former means that the turker agrees with the prediction of the network, while the latter means that the example is misclassified. We take the following measures to ensured high quality answers: (1) we only recruit \emph{master turkers}, \textit{i.e.} turkers who have demonstrated excellence across a wide range of tasks and are awarded the masters qualification by Amazon (2) we mitigate the effect of bad turkers by aggregating 5 independent answers for every question (3) we manually sample more than 500 questions and successfully cross-validated our own guesses with the answers from the turkers, finding that 4 or more positive answers over 5 leads to almost no false positives.

\begin{figure}[!htb]
\centering
\includegraphics[width=.5\linewidth]{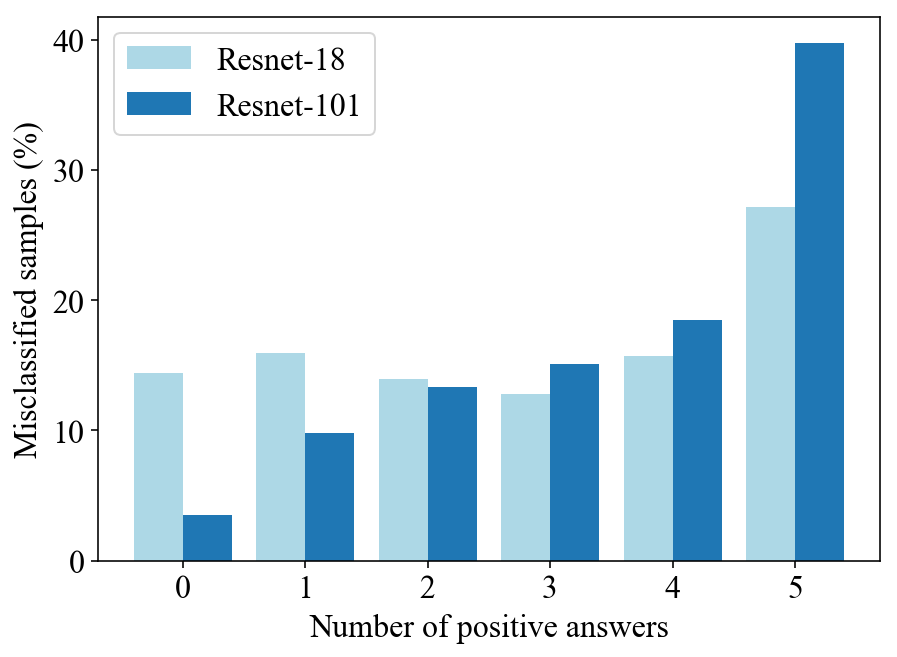}
\caption{Positive answers for misclassified samples. An image is prompted to 5 different subjects and a \textbf{positive} answer means the subject \textbf{agrees with the predicted class}}
\label{hist_resnet}
\end{figure}

Figure~\ref{hist_resnet} shows a breakdown of the misclassified images by a ResNet-101 (resp. a ResNet-18) according to the number of positive answers they receive from the turkers. Our first results show that for a ResNet-101 (resp. a ResNet-18), for $39.76\%$ (resp. $27.14\%$) of the misclassified examples, all the turkers $(5/5)$ agree with the prediction of the model. If we further consider the prediction of the model for a given image to be correct if at least four turkers out of five $(4/5)$ agree with it, the rectified Top-1 error of the models are drastically reduced.

Table~\ref{table_misclassifications} shows the original Top-1 errors together with the rectified versions. For the ResNet-101 and the DenseNet-161, the rectified Top-1 error is respectively $9.47\%$ and $10.87\%$. Similarly, the top-5 error of the ResNet-101 on a single crop is $6.44\%$. When we present the misclassified images to the turkers together with the top 5 predictions of the model, the rectified top-5 error drops to $1.94\%$. When instead of submitting to the turker the misclassified images (\textit{i.e} top-1-misclassified images), we present them the pictures for which the ground truth is not in the top-5 predictions (\textit{i.e.} top-5-misclassified images), the rectified top-5 error drops further to $1.37\%$. This shows that while the top-5 score is often used to mitigate the fact that many classes are present in the same image, it does not reflect the whole multi-label nature of those images. Moreover, the observation on top-5 is in line with the conclusions regarding top-1. If top-5 is the important measure, this experiment suggests that ImageNet is (almost) solved as far as accuracy is concerned, therefore explaining the marked slowdown in performance improvement observed recently on ImageNet.
%These performances are significantly lower than the original Top-1 error for both models and leave less room for improvement than what one would expect, therefore explaining the marked slowdown in performance improvement observed recently on ImageNet. %|Similarly, the the ResNet101, the rectified top-5 error is $1.94\%$ instead of $6.44\%$

\begin{table}[!htb]
\centering
\vspace{10pt}
\caption{Standard and rectified Top-1 errors for various models. Rectified Top-1 errors are significantly lower.}
\begin{tabular}{l|c|c}
\hline
Model & Top-1 error & Rectified Top-1 error\\
\hline
Resnet-18    & 31.39 & 17.93 \\
Resnet-101   & 22.69 &  9.47 \\
Densenet-121 & 25.53 & 14.37 \\
Densenet-161 & 22.85 & 10.87 \\
\hline
\end{tabular}
\label{table_misclassifications}
\end{table}

\begin{figure}[!htb]
\centering
\includegraphics[width=\linewidth]{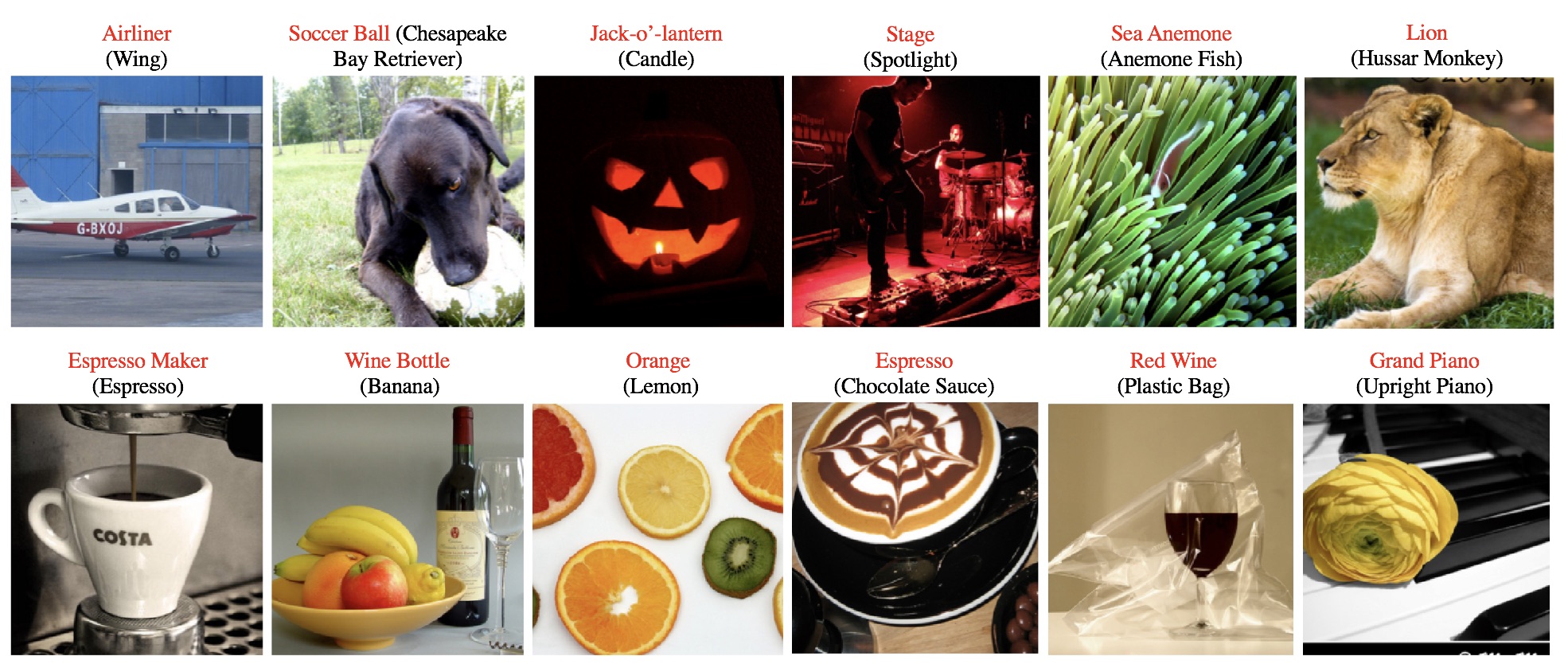}
   \caption{Some test samples misclassified by a ResNet-101 (first row) and a Densenet-161 (second row). The predicted class is indicated in {\color{red}{red}}, the ground truth in {black} and in parenthesis. All those examples gathered more than four (4 or 5) positive answers over 5 on AMT. Note that no adversarial noise has been added to the images.}
\label{fig:short}
\label{img_misclassifications}
\end{figure}

The difference between the ground truth labeling of the images and the predictions of the models validated by the turkers can be traced back to the collection protocol of ImageNet. Indeed to create the dataset, Deng et al.~\cite{deng2009imagenet} first queried several image search engines (using the synsets extracted from WordNet) to obtain good candidate images. Turkers subsequently cleaned the collected images by validating that each one contains objects of a given synset. This labeling procedure ignores the \emph{intrinsic multilabel} nature of the images, neither does it take into account important factors such as \emph{composionality}. Indeed, it is natural that the label \texttt{wing} is also relevant for a picture displaying a plane and labelled as \texttt{airliner}. Figure~\ref{img_misclassifications} shows examples of misclassifications by a RestNet-101 and a DenseNet-161. In most cases, the predicted class is (also) present in the image.
%See Fig. \ref{hist_resnet} and \ref{img_misclassifications}

% \begin{figure}
% \begin{subfigure}{0.5\textwidth}
% \includegraphics[width=\linewidth]{hist_resnet.png}
% \caption{}
% \label{hist_resnet}
% \end{subfigure}
% \hspace*{\fill} % separation between the subfigures
% \begin{subfigure}{0.5\textwidth}
% \vspace{41pt}
% \begin{tabular}{l|c|c}
% \hline
% Model & Top-1 error & Rectified Top-1 error\\
% \hline
% Resnet-18    & 31.39 & 17.93 \\
% Resnet-101   & 22.69 &  9.47 \\
% Densenet-121 & 25.53 & 14.37 \\
% Densenet-161 & 22.85 & 10.87 \\
% \hline
% \end{tabular}
% \caption{}
% \vspace{14pt}
% \label{table_misclassifications}
% \end{subfigure}
% \caption{(a) \small Positive answers for misclassified samples. An image is prompted to 5 different subjects and a \textbf{positive} answer means the subject \textbf{agrees with the predicted class}. (b) Standard and rectified Top-1 errors for various models. Rectified Top-1 errors are significantly lower.}
% \end{figure}

\subsection{Study of Robustness to Adversarial Examples}

We conducted a human subject study to investigate the appreciation of adversarial perturbation by end-users. Similarly to the human subject study of misclassifications, we used the center-cropped images of the ImageNet validation set. Next, we consider a subset of 20 classes and generate an adversarial example from each legitimate image using the IFGSM attack on a pre-trained ResNet-101. We used a step size of $\epsilon = 0.001$ and a maximum number of iterations of $M=10$. This attack deteriorated the accuracy of the network on the validation set from $77.31\%$ down to $7.96\%$. Note that by definition, we only need to generate adversarial samples for the correctly predicted test images. We also only consider non-targeted attacks since targetted attacks often require larger distorsion from the adversary and are more challenging if the target categories are far from the ones predicted in case of non-targetted attacks.

We consider two settings in this experiment. In the first configuration, we present the turkers with the whole adversarial image together with the prediction of the network. We then ask each turker if the predicted label is relevant to the given picture. Again, the possible answers are \emph{yes} and \emph{no}. Each image is shown to five (5) different turkers. In the second configuration, we show each turker the interpretation image generated by LIME instead of the whole adversarial image using the top 8 most important features. The rest of the experimental setup is identical to the previous configuration where we showed the whole images to the turkers. If a turker participated in the second experiment with the interpretation images after participating in the first experiment where the whole images are displayed, his answers could be biased. To avoid this issue, we perform the two studies with three days intervals. Similarly to our previous experiments, we report the rectified Top-1 error on adversarial examples by considering the prediction of the model as correct if at least $4/5$ turkers agree with it.

\begin{figure}[!htb]
\centering
\includegraphics[width=.5\linewidth]{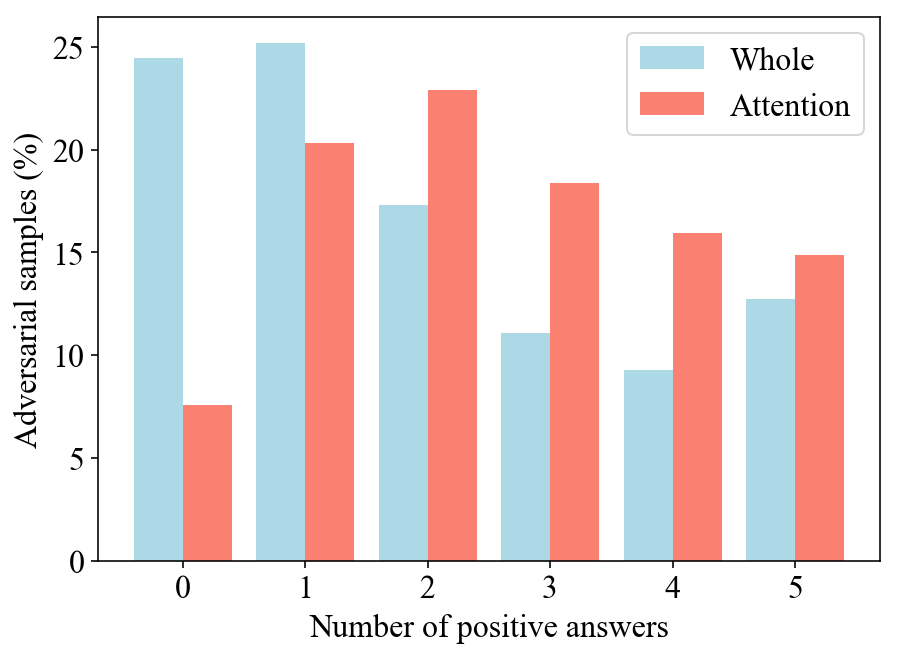}
\caption{Positive answers for adversarial samples. The images are either displayed as a whole (entire image) or with attention (explanation). Every image is prompted to 5 different subjects. A \textbf{positive} answer means the subject \textbf{agrees with the predicted adversarial class}.}
\label{hist_attention}
\end{figure}

% \begin{figure}[!htb]
% \centering
%    \begin{minipage}{0.48\textwidth}
% \centering
%   \includegraphics[width=\linewidth]{hist_attention.png}
%   \caption{Positive answers for adversarial samples. The images are either displayed as a whole (entire image) or with attention (explanation). Every image is prompted to 5 different subjects. A \textbf{positive} answer means the subject \textbf{agrees with the predicted adversarial class}.}
% \label{hist_attention}
%    \end{minipage}
%    \begin {minipage}{0.48\textwidth}
% \vspace{41pt}
% \begin{tabular}{l|c|c}
% \hline
% Setting & Top-1 error & Rectified Top-1 error\\
% \hline
% Whole image & 92.04 & 76.78 \\
% Explanation   & 92.04 & 70.68  \\
% \hline
% \end{tabular}
% \vspace{14pt}
% \caption{Standard and rectified Top-1 errors on adversarial examples for the whole images and the explanations.}
% \label{adv_errors}
% \end{minipage}
% \end{figure}

\begin{figure}[!htb]
   \centering
   \includegraphics[width=0.5\linewidth]{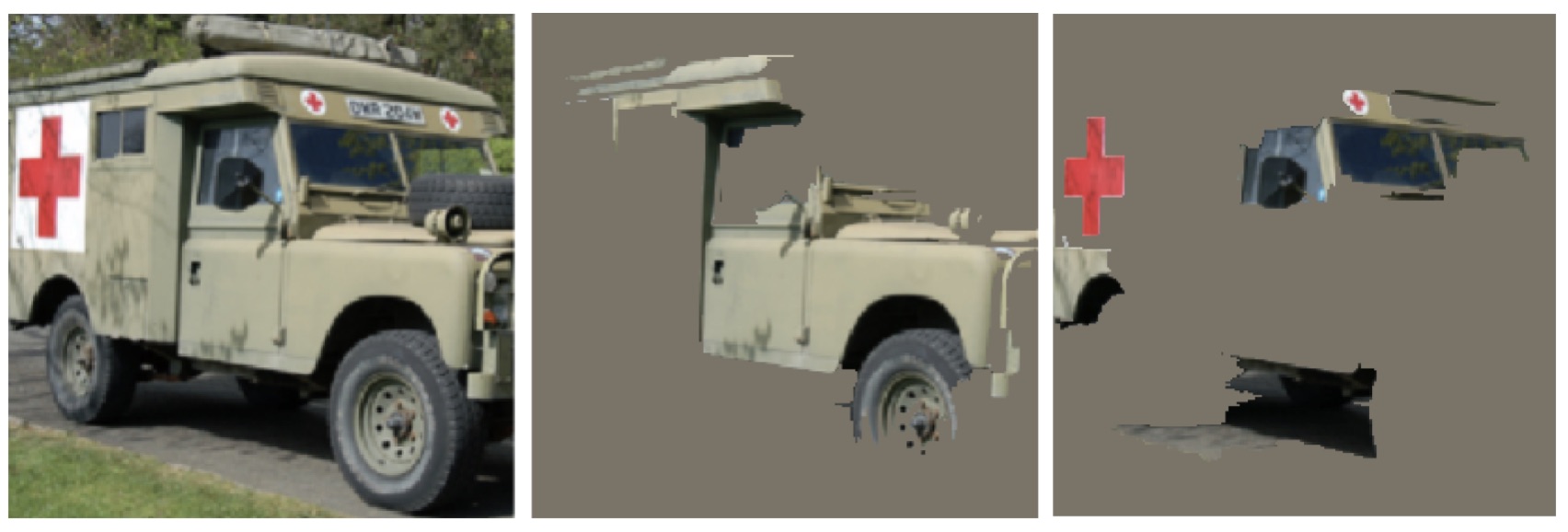}
   \caption{\textbf{Left:} an adversarial image of the true class \texttt{Jeep} predicted as \texttt{Ambulance} by the network. \textbf{Center:} the explanation of the clean image for its prediction (\texttt{Jeep}). \textbf{Right:} the explanation of the adversarial image for its prediction  (\texttt{Ambulance}).}
\label{fig:short}
\label{jeep}
\end{figure}

Table~\ref{adv_errors} shows the standard and the rectified Top-1 errors for the adversarial examples. Two observations can be made. First, the robustness of the models as measured by the Top-1 error on adversarial samples generated from the validation set of ImageNet is also underestimated. Indeed, when the whole images are displayed to them, the turkers agree with  $22.01\%$ of the predictions of the networks on adversarial examples. This suggests that often, the predicted label is easily identifiable in the picture by a human. Figure~\ref{jeep} shows an example (labeled \texttt{jeep}) with the explanations of the predictions for the legitimate and adversarial versions respectively. One can see that the adversarial perturbation exploits the ambiguity of the image by shifting the attention of the model towards regions supporting the adversarial prediction (\textit{e.g.} the red cross).

\begin{table}
\centering
\caption{Standard and rectified Top-1 errors on adversarial examples for the whole images and the explanations.}
\vspace{10pt}
\begin{tabular}{l|c|c}
\hline
Setting & Top-1 error & Rectified Top-1 error\\
\hline
Whole image & 92.04 & 76.78 \\
Explanation   & 92.04 & 70.68  \\
\hline
\end{tabular}
\label{adv_errors}
\end{table}

\begin{figure}[!htb]
   \centering
   \includegraphics[width=\linewidth]{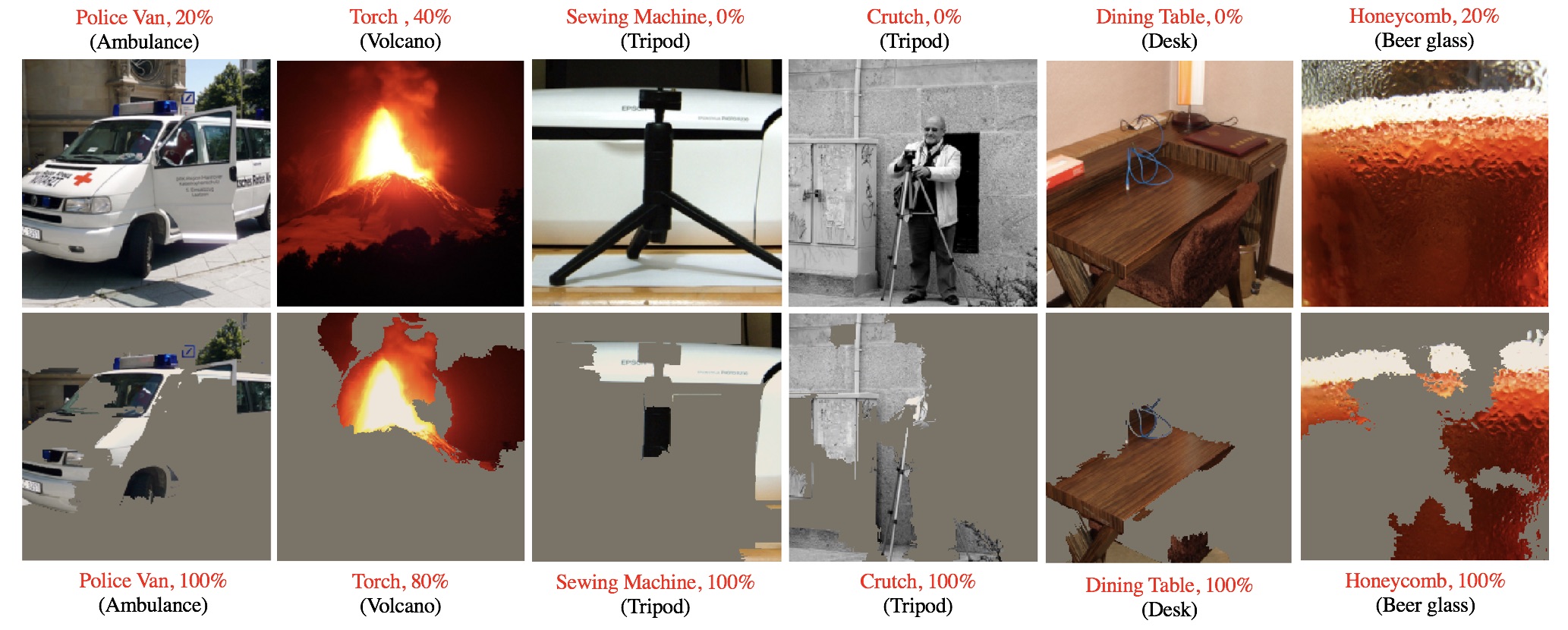}
   \caption{Adversarial samples, displayed as a whole (entire image, first row) or with explanation (only $k$ top super-pixels, second row). The adversarial class is indicated in {\color{red}{red}}, the true class in black and between parenthesis. In {\color{red}{red}} is also displayed the percentage of positive answers for the displayed image (over 5 answers). In both cases, a \textbf{positive} answer means the subject \textbf{agrees with the predicted adversarial class}.}
\label{fig:long}
\label{fig:onecol}
\label{samples_attention}
\end{figure}

The second substantial observation is that the percentage of agreement between the predictions of the model and the turkers increases from  $22.01\%$ to $30.80\%$ when the explanation is shown instead of the whole image (see also Figure~\ref{hist_attention}). We further inspected the adversarial images on which the turkers agree with the model's prediction when the explanation is shown but mostly disagree when the whole image is shown. Figure~\ref{samples_attention} displays examples of such images. In most cases, even though the predicted label does not seem correct when looking at the whole image, the explanation has one of the two following effects. It either reveals the relevant object supporting the prediction (e.g., \texttt{Tripod} image) or creates an ambiguous context that renders the predicted label plausible (e.g., \texttt{Torch} or \texttt{Honeycomb} images). In all cases, providing explanations to the user mitigates the impact of misclassifications due to adversarial perturbations.

%\paragraph{Remark} \red{Note that similar results hold for other non-targeted attacks such as DeepFool \cite{moosavi2015deepfool}, \textit{i.e.} the results presented above are independent of the non-targeted attack. We did not consider targeted attacks because such attacks require much more distortion (!!Reference!!) (as measured by $L2$ norm for example) of the image to be effective, hence the popularity of non-targeted attacks which are more subtle and difficult to counteract.}

\subsection{Adversarial Examples for Model Criticism}

Example-based explanation methods such as model criticism aim at summarizing the statistics learned by a model by using a carefully selected subset of examples. MMD-critic proposes an effective selection procedure combining prototypes and criticisms for improving interpretability. Though applicable to a pre-trained neural network by using the hidden representation of the examples, it only indirectly exploits the discriminative nature of the classifier. In this work, we argue that adversarial examples can be an accurate alternative to MMD-critic for model criticism. In particular, they offer a natural way of selecting prototypes and criticism based on the number of steps of FGSM necessary to change the decision of a classifier for a given example. Indeed, for a given class and a fixed number of maximum steps $M$ of IFGSM, the examples that are still correctly classified after $M$ steps can be considered as useful prototypes since they are very representative of what the classifier has learned from the data for this class. In contrast, the examples whose decision change after $1$ or few steps of FGSM are more likely to be valid criticisms because they do not quite fit the model.

We conducted a human study to evaluate our hypothesis.
At each round, we present a turker with six classes represented by six images each, as well a target sample randomly drawn from one of those six classes. We measure how well the subject can assign the target sample to the class it belongs. The assignment task requires a class to be well-explained by its six images. The challenge is, therefore, to select well those candidate images. The adversarial method selects as prototypes the examples not misclassified IFGSM after $M=10$ steps, and as criticsms the examples misclassified after one step.
In addition to MMD-critic (using $\lambda = 10^{-5}$), we compare the adversarial approach to a simple baseline using the probabilities for selecting the prototypes (considerable confidence, e.g., $>0.9$) and the criticisms (little confidence, e.g., $<0.1$). We also compare with the baseline randomly sampling examples from a given class. For each method (except the random baseline), we experiment with showing six prototypes only vs. showing three prototypes and three criticisms instead.

To properly exploit the results, we discarded answers based on the time spent by the turkers on the question to only consider answers given in the range from 20 seconds to 3 minutes. This resulted in $1,600$ valid answers for which we report the results in Table~\ref{protos_results}. Two observations arise from those results. First, the prototypes and criticisms sampled using the adversarial method represent better the class distribution and achieve higher scores for the assignment task both when only prototypes are used ($55.02\%$) and when prototypes are combined with criticisms ($57.06\%$). Second, the use of criticisms additionally to the prototypes always helps to better grasp the class distribution.

\begin{figure}[!h]
\begin{subfigure}{0.5\textwidth}
\includegraphics[width=\linewidth]{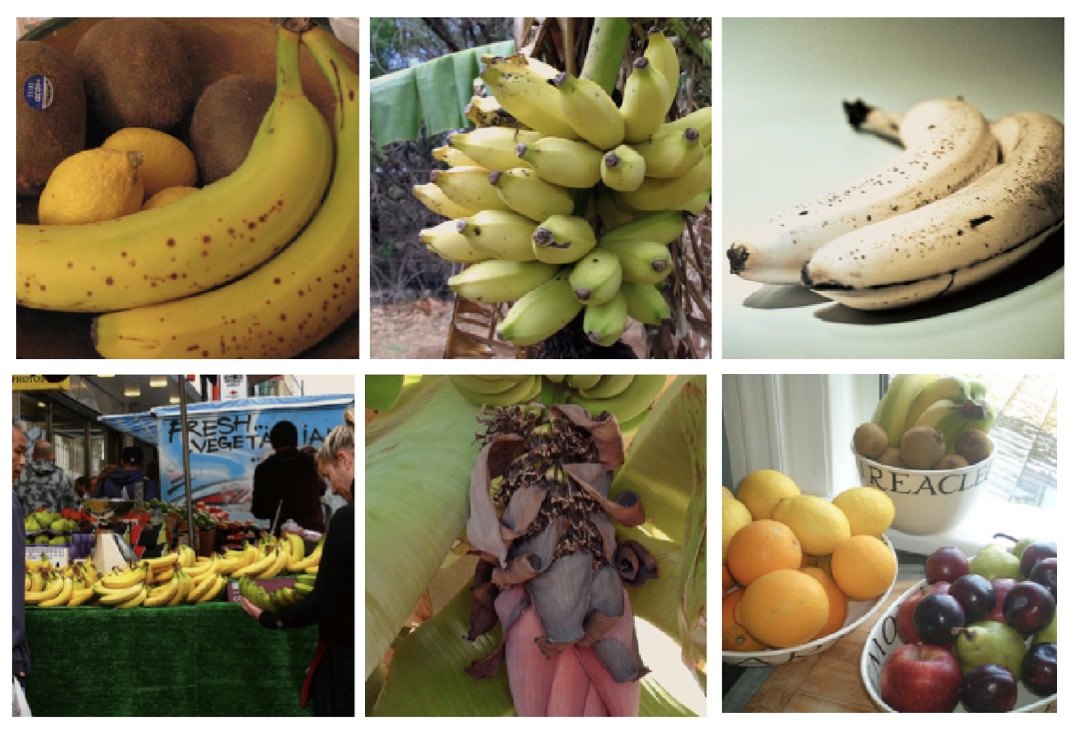}
\caption{}
\label{banana_mmd}
\end{subfigure}
\hspace*{\fill} % separation between the subfigures
\begin{subfigure}{0.5\textwidth}
\includegraphics[width=.98\linewidth]{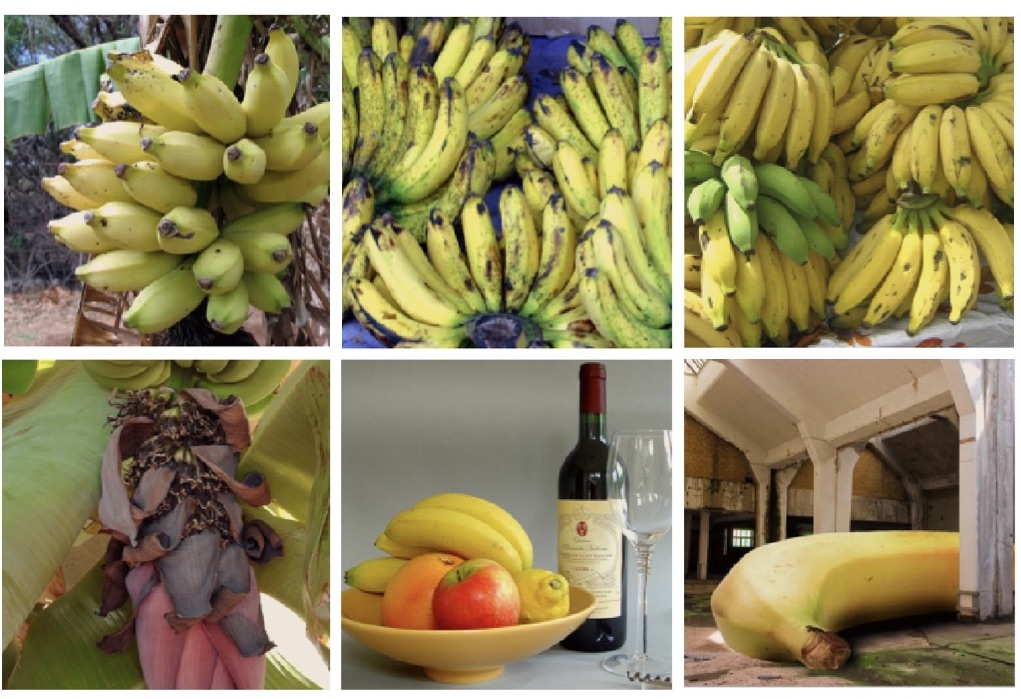}
\caption{}
\label{banana_adv}
\end{subfigure}
\vspace{-10pt}
\caption{(a) Prototypes (first row) and Criticisms (second row) for the \texttt{Banana} class when using the MMD-critic method. (b) Prototypes (first row) and Criticisms (second row), for the \texttt{Banana} class when using the adversarial selection.}
\end{figure}

More qualitatively, we display in Figures~\ref{banana_mmd} and~\ref{banana_adv} some prototypes and criticisms for the class \texttt{Banana} generated using respectively the MMD and the adversarial method for the test samples, that demonstrate the superiority of the adversarial method.

\begin{table}[!htb]
\caption{Results of the Prototype and Criticism study. We report the average success rate of the task (assign a target image to one of the six groups) for the seven conditions. The adversarial selection procedures outperforms MMD-critic as well as the other baselines used in the study.}
\label{protos_results}
\centering
\vspace{10pt}
\begin{tabular}{l|c|c}
\hline
Condition & Mean (\%)& Std (\%)\\
\hline
Adv - Prototypes and Criticisms   & \bf 57.06 & 3.22\\
MMD - Prototypes and Criticisms   & 50.64 & 3.25\\
Probs - Prototypes and Criticisms & 49.36 & 3.26\\
Adv - Prototypes only             & 55.02 & 3.23\\
MMD - Prototypes only             & 53.18 & 3.24\\
Probs - Prototypes only           & 52.44 & 3.25\\
Random                            & 49.80 & 3.25\\
\hline
\end{tabular}
\end{table}

\subsection{Uncovering Biases with Model Criticism}

To uncover the undesirable biases learned by the model, we use our adversarial example approach to model criticism since it worked better than MMD-critic in our previous experiments.
We consider the class \texttt{basketball} (for which humans often appear in the images). We select and inspect a reduced subset of prototypes and criticisms from the category \texttt{basketball}.
%\newpage (validation)
The percentage of \texttt{basketball} training images on which at least one white person appears is about $55\%$. Similarly, the percentage of images on which at least one black person appears is $53\%$. This relative balance contrasts with the statistics captured by the model. Indeed, on the one hand, we found that about $78\%$ of the prototypes contain at least one black person and only $44\%$ for prototypes contain one white person or more. On the other hand, for criticisms, $90\%$ of the images contain at least one white person and only about $20\%$ include one black person or more. This suggests that the model has learned a biased representation of the class \texttt{basketball} where images containing black persons are prototypical. To further validate this hypothesis, we sample pairs of similar pictures from the Internet.

\newpage
\begin{figure}[!h]
   \centering
   \includegraphics[width=\linewidth]{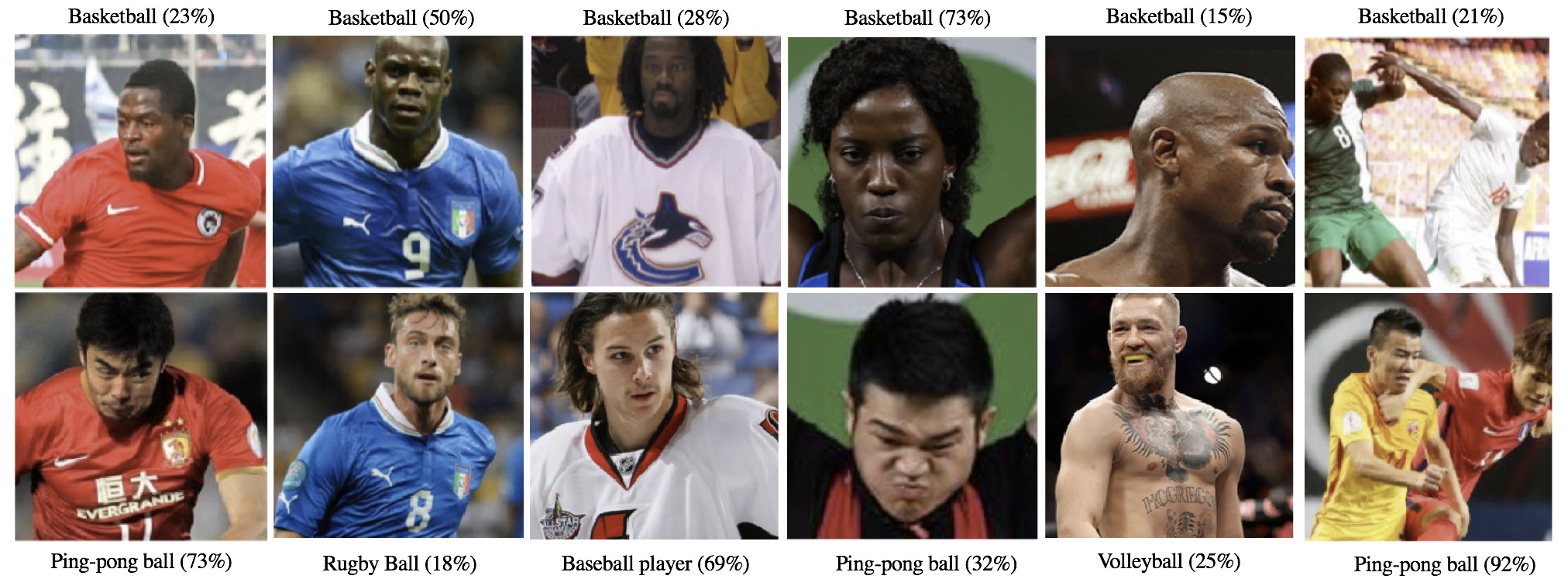}
   \caption{Pairs of pictures (columns) sampled over the Internet along with their prediction by a ResNet-101.}
   \label{basket_comp}
\end{figure}
\begin{figure}[!h]
\vspace*{-1.2cm}
   \centering
   \includegraphics[width=\linewidth]{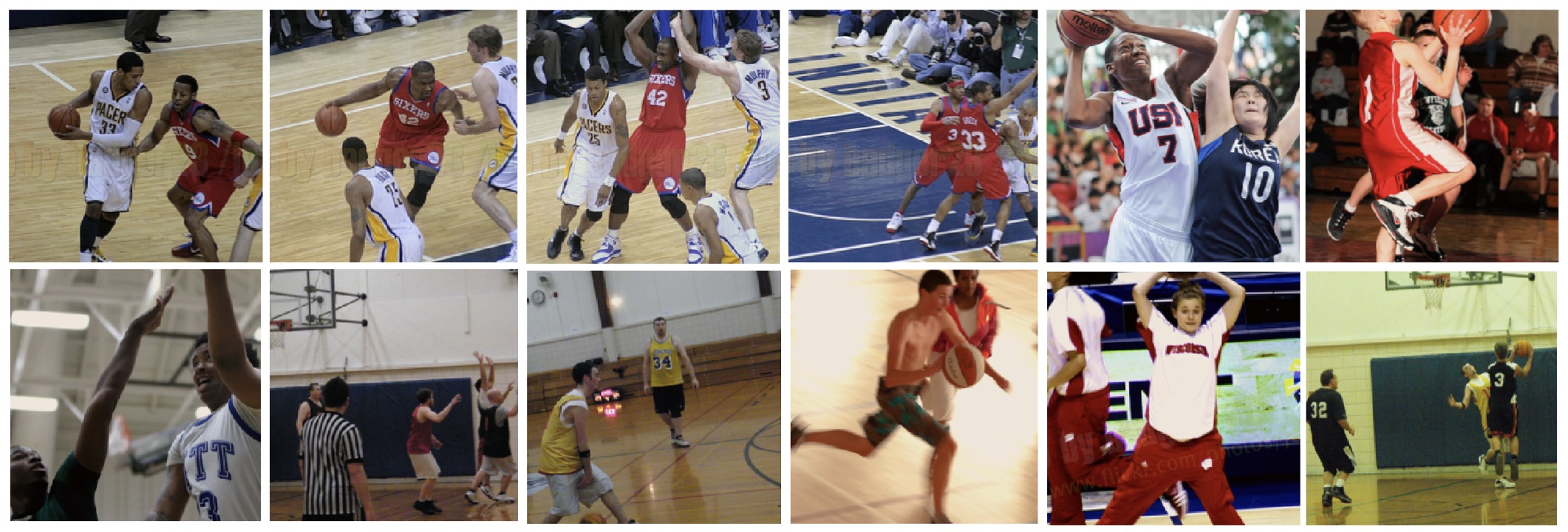}
   \caption{Prototypes (first row) and Criticisms (second row) for the test images of the \texttt{Basketball} class, adv method.}
   \label{protos_bb}
\end{figure}
\begin{figure}[!h]
\vspace*{-1.2cm}
\begin{subfigure}{0.5\textwidth}
\includegraphics[width=\linewidth]{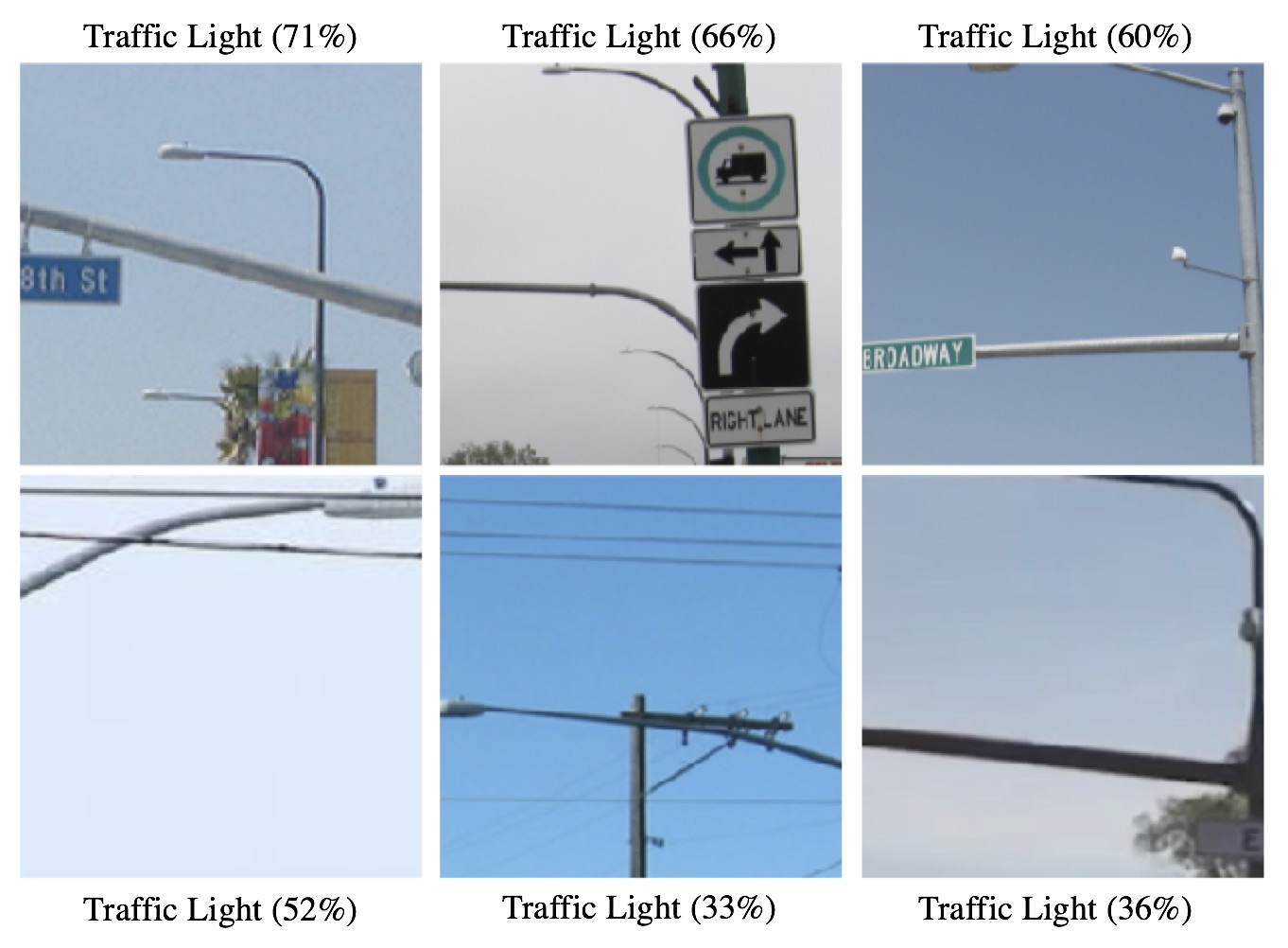}
\caption{}
\label{bias_tl}
\end{subfigure}
\hspace*{\fill} % separation between the subfigures
\begin{subfigure}{0.5\textwidth}
\includegraphics[width=\linewidth]{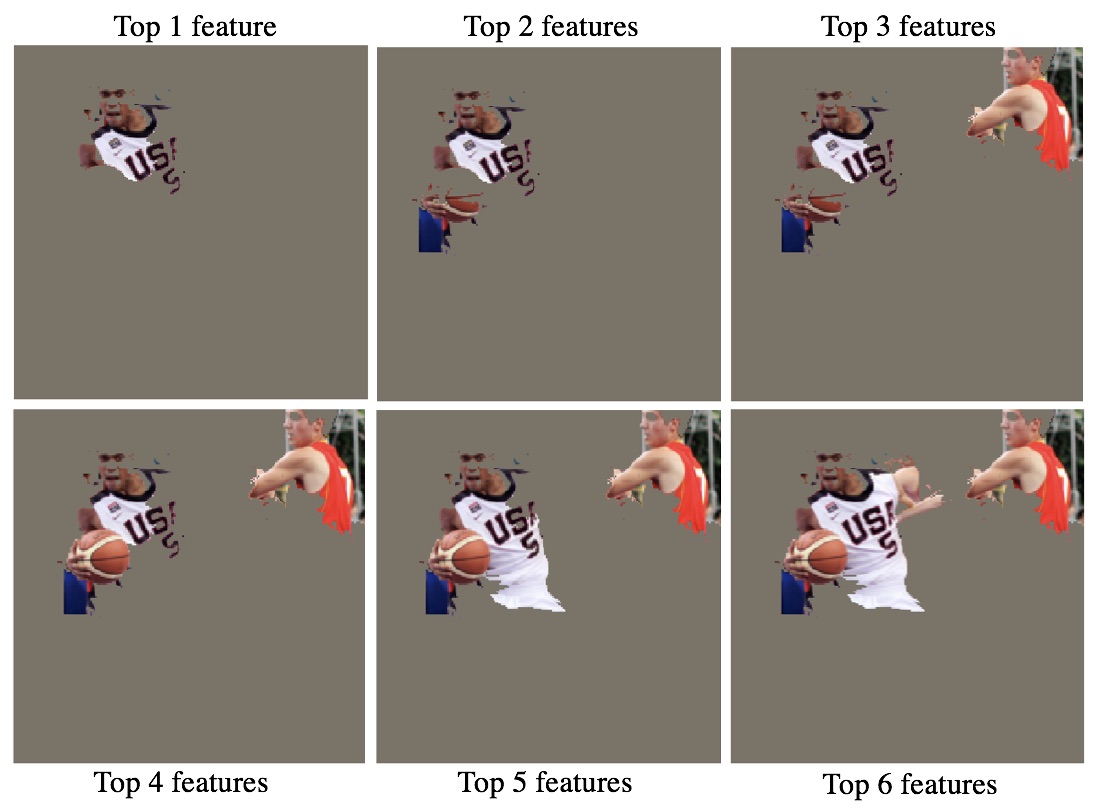}
\caption{}\label{features_bb}
\end{subfigure}
\vspace{-10pt}
\caption{(a) Images from the Internet classified as \texttt{Traffic Light} for a ResNet-101. (b) Explanation path of a ResNet-101 for a test sample of the \texttt{Basketball} class. Note that the first feature to appear includes the writing on the jersey and then the basket ball.}
\end{figure}

The pairs are sampled such that the primary apparent difference between the two images is the skin color of the persons. We then fed images to the model and gathered the predictions. Figure~\ref{basket_comp} shows the results of the experiments. All images containing a black person are classified as basketball while similar photos with persons of different skin color are labeled differently. Figure~\ref{features_bb} provides additional supporting evidence for this hypothesis. It shows the progressive feature based explanation path uncovering the super-pixels of an example (correctly classified as \texttt{basketball}) by their order of importance. The most importance super-pixels depict the jersey  and the skin color of the player. The reasons why the model learns these biases are unclear. One hypothesis is that despite the balanced distribution of races in pictures labeled basketball, black persons are more represented in this class in comparison to the other classes. A similar phenomenon has also~ been noted in the context of textual data.
\cite{paperno2014corpus,bolukbasi2016man}. We defer further investigations on this to future studies.
\paragraph{Remark} We have focused on racial biases because a human can easily spot them. In fact, we have found similar biases where pictures displaying Asians dressed in red are very often classified as \texttt{ping-pong ball}. However, we hypothesize the biases of the model are numerous and diverse. For example, we also have found that the model often predicts the class \texttt{traffic light} for images of a blue sky with street lamps as depicted in Figure~\ref{bias_tl}. In any case, model criticism has proven effective in uncovering the undesirable hidden biases learned by the model.

\section{Conclusion}

Through human studies and explanations, we have proven that the performance of SOTA models on ImageNet is underestimated. This leaves little room for improvement and calls for new large-scale benchmarks involving for example multi-label annotations. We have also improved our understanding of adversarial examples from the perspective of the end-user and positioned model criticism as a valuable tool for uncovering undesirable biases. These results open an exciting perspective on designing explanations and automating bias detection in vision models. Our study suggests that more research in these topics will be necessary to sustain the use of machine learning as a general purpose technology and to achieve new breakthroughs in image classification.

%\begin{table}
%\centering
%\begin{tabular}{l|c|c|c}
%\texttt{Basketball} & White (\%) & Black (\%) & None (\%) \\
%\hline
%All             &  55.09 & 52.82 & 8.48 \\
%Test Prototypes &  44.45 & 77.78 & 0.00 \\
%Test Criticisms &  90.00 & 20.00 & 0.00 \\
%\hline
%\end{tabular}
%\label{bias_racial_bb}
%\caption{Proportion of samples in the \texttt{Basketball} class displaying at least of person of the mentioned skin color (black or white). Note that the image can only display a basket ball, in which case it is counted in \textit{None}. The prototypes are defined here as the test images requiring more than 8 steps of the IFGS method to fool a ResNet-101 (9 in total over 50 test samples), and the criticism those requiring only 1 or 2 steps (10 in total over 50 test samples).}
%\end{table}
%Note that the first feature to appear includes the writing on the jersey and then the basket ball, along with another player.

\bibliographystyle{splncs}
\bibliography{egbib}
\end{document}

%% file: macros.tex
\newcommand{\bx}{x}%\xvector{x}}
\newcommand{\by}{y}
\newcommand{\bxprime}{\tilde{x}}%\xvector{\tilde{x}}}
\newcommand{\network}{g}

\newcommand{\loss}{\ell}

\newcommand{\adv}[1]{\tilde{#1}}
\newcommand{\perturb}[1]{\delta_{#1}}
\newcommand{\adveps}{\epsilon}